\documentclass{ecai}
\usepackage{times}
\usepackage{graphicx}
\usepackage{latexsym}
\usepackage{xcolor}
\usepackage{amsmath,amssymb}
\usepackage{stmaryrd}

\newcommand{\KL}{D_{\mathrm{KL}}}

\newcommand{\citet}[1]{\citeauthor{#1} \shortcite{#1}}
\usepackage{booktabs}
\usepackage[tight,footnotesize]{subfigure}

\begin{document}

\title{Improving Unsupervised Domain Adaptation with Variational Information Bottleneck}

\author{
Yuxuan Song\institute{Shanghai Jiao Tong University, email: songyuxuan@apex.sjtu.edu.cn}~~,
Lantao Yu\institute{Stanford University, email: lantaoyu@cs.stanford.edu}~~,
Zhangjie Cao$^2$,
Zhiming Zhou$^1$\\
Jian Shen$^1$,
Shuo Shao$^1$,
Weinan Zhang$^1$,
Yong Yu$^1$
}

\maketitle
\bibliographystyle{ecai}

\begin{abstract}
Domain adaptation aims to leverage the supervision signal of source domain to obtain an accurate model for target domain, where the labels are not available. To leverage and adapt the label information from source domain, most existing methods employ a feature extracting function and match the marginal distributions of source and target domains in a shared feature space. In this paper, from the perspective of information theory,
we show that representation matching is actually an \textit{insufficient} constraint on the feature space for obtaining a model with good generalization performance in target domain. We then propose variational bottleneck domain adaptation (VBDA), a new domain adaptation method which improves feature transferability by explicitly enforcing the feature extractor to ignore the task-irrelevant factors and focus on the information that is essential to the task of interest for both source and target domains.
Extensive experimental results demonstrate that VBDA significantly outperforms state-of-the-art methods across three domain adaptation benchmark datasets.
\end{abstract}

\section{Introduction}
Deep learning has shown impressive abilities
on solving numerous machine learning tasks.  
Most of recent advances heavily rely on the access to huge amount of labeled data and the assumption that both training and test data are sampled from the same underlying distribution.
However,
there are many application scenarios where the labeled data for the task of interest (target domain) is hard to obtain, while another correlated domain (source domain) with non-negligible dissimilarity consists of sufficient annotated data. 
Hence, there is strong motivation to leverage the supervision signal from source domain to help build an effective model in target domain. 
Learning an accurate predictive model for target domain with the presence of covariate shift \cite{sugiyama2012machine} (\emph{i.e.}, the input data distributions of source and target domains are different) is known as domain adaptation.
In this paper, we focus on a general and challenging setting where no label information is available in target domain, which is termed as unsupervised domain adaption.

Recent advances in deep learning stimulate a fruitful line of domain adaptation works, which leverage deep neural networks to infer the latent variables and match the marginal distributions of source and target domains in the latent space \cite{ganin2016domain,long2015learning,RTN}.
Inspired by Generative Adversarial Networks \cite{goodfellow2014generative}, an adversarial domain adaptation mechanism is utilized in \cite{ganin2016domain,tzeng2017adversarial,luo2017label,xie2018learning}. This mechanism involves a two-player game between a discriminator and a feature extractor: the domain discriminator is trained to tell whether the samples come from source or target domain, while the feature extractor is trained to maximize the discriminator's classification error. Essentially, adversarial domain adaptation methods seek to minimize the Jensen-Shannon divergence between source and target distribution of latent features.

However, it has been shown that matching the marginal distribution in latent feature space is not strong enough for ensuring the essential information to be transferred \cite{zhao2019learning}.
It is possible that the learned mapping is misled by the domain invariant yet task-irrelevant factors and fails to capture the semantic information. Consider the following example, the adaptation task is to recognize animals in the pictures, and in source domain, most of the sheep are appearing with the grassland as the background and most of the horses are appearing with the animal house as the background; while in target domain the background configuration of the two species are random and marginal distributions of the background are the same.  In cases like this, directly matching the marginal feature distributions could result in that the domain-invariant yet task-irrelevant information, (\emph{i.e.} the background in the above example), outweigh the task-relevant information, which then lead to worse performance on target domain (also known as negative transfer \cite{pan2010survey}).

To tackle the lack of semantic alignment, many recent works proposed to enhance the label information of target domain based on some strong assumptions \cite{luo2017label,shu2018dirt,xie2018learning,saito2017asymmetric}. 
One of the most widely used hypotheses is the cluster assumption \cite{grandvalet2005semi} (also known as low density separation assumption), which states that the data instances are distributed into several separate clusters and samples in the same cluster share the same label. However, the cluster assumption is actually too strong and inappropriate for many practical scenarios, and directly using the cluster assumption could bring non-negligible undesired effects \cite{shu2018dirt}.

In this paper, inspired by the information bottleneck principle, we propose a simple yet effective regularization technique for domain adaptation methods by combining conditional entropy minimization and variational information bottleneck, which enforces the feature extractor to ignore the irrelevant factors and focus on the essential information for the task of interest (\textit{i.e.}, the sufficient statistics for determining the parameters of the predictive models). Our method tends to learn a balanced and clean representation space (\textit{i.e.}, no information preference on source or target domain and less irrelevant factors), which improves the generalization ability of the predictive model and renders strong yet widely used assumptions such as the cluster assumption more realistic. We further provide a theoretical analysis on the generalization error bound in Section~\ref{domain adapt theory}. Extensive experimental results demonstrate that our model outperforms state-of-the-art methods across three domain adaptation benchmark datasets \cite{officehome,office31,svhn}.

\section{Related Works}

In this section, we discuss several most relevant works in the field of domain adaptation. \cite{ganin2016domain} and \cite{long2015learning} 
proposed to project the source and target domain into a common representation space, and encouraged the corresponding marginal feature distribution to be matched under the guidance of some distance or divergence. Adversarial techniques based on the framework of GAN \cite{goodfellow2014generative} are widely explored in the literatures of domain adaptation \cite{saito2017asymmetric11,hoffman2017cycada,tzeng2017adversarial}, which corresponds to minimizing the symmetric Jensen-Shannon divergence. 
However, \cite{JAN} 
pointed out that adversarial domain adaptation methods which only match the marginal distribution are problematic and insufficient for successful adaptation.
To address this limitation, various methods have been proposed. For example, \cite{JAN,CDAN} proposed to match
the joint distribution instead of purely matching the marginal; \cite{ghifary2016deep} introduced a decoder architecture for capturing the semantic information; and \cite{hoffman2017cycada} utilized cycle consistency constraints to preserve semantic information.
However, the main limitation of these methods is that, although the semantic information is enhanced, the learned representation is still likely to preserve  domain-invariant factors that are \emph{irrelevant} to the predictive task, which may mislead the semantic alignment especially when training samples are not sufficient enough. In the animal recognition example mentioned in Introduction section,
the background is the domain-invariant yet \emph{irrelevant} factors. The learned representation tended to preserve the background information due to the fact that the background has statistically dependency with the class label in source domain and the marginal distribution of background is invariant between the source and target domain. And irrelevant information will disturb the predictive task on target domain.
Hence there is strong motivation to enforce the feature extractor to only focus on the essential information for the task of interest and ignore  as much irrelevant factors as possible, no matter they are domain-invariant or not. Inspired by this intuition, we propose to regularize domain adaptation models with information bottleneck principle \cite{tishby2000information}, which seeks to find the optimal tradeoff between representation accuracy and compression. Since information bottleneck method has been successfully applied to supervised learning \cite{alemi2016deep}, generative modeling \cite{jeon2018ib,peng2018variational} and reinforcement learning \cite{peng2018variational}, in the context of domain adaptation, we propose to exploit it to preserve sufficient statistics and remove irrelevant factors in the learned representations. While \cite{motiian2016information} also augment domain adaptation with information bottleneck, they focus on a specific scenario, where an auxiliary data view (\textit{e.g.}, skeleton data for gestures and bounding box for objects) is available and the information bottleneck is incoporated to leverage these additional data view. In contrast, our method seeks to provide a new regularization technique for general unsupervised domain adaptation with deep neural networks.

On the other hand, to counter the lack of attention on target semantic information, conditional entropy minimization \cite{grandvalet2005semi} is widely used in unsupervised domain adaptation \cite{RTN,luo2017label}.
These methods are based on the cluster assumption that, the decision boundary should not cross high density regions, but instead lie in low density regions \cite{chapelle2005semi}. In other words, it assumes that the data instances are distributed into several separate clusters, and samples in the same cluster share the same class label. However, it should be noted that the cluster assumption can be too strong to be satisfied in many practical scenarios, which will bring undesired effects to the stability of training and performance of the models. Essentially, the cluster assumption in the representation space is satisfied only when the learned representations merely preserve semantic information that is relevant to the predictive task, while our variational bottleneck domain adaptation framework intrinsically seeks to find such a clean representation space  which renders the cluster assumption more realistic and achieves better feature transferability.

\section{Background \& Notations}\label{sec:preliminary}
\subsection{Domain Adaptation}
To describe a domain, we introduce a joint data distribution $p(x,y)$ with which we define both the marginals and conditionals. Let $p_s(x_s, y_s)$ denote the underlying joint data distribution of the data instance $x_s$ and the corresponding label $y_s$ for source domain, and let $p_s(x_s)$ denote the marginal distribution of $x_s$. $p_t(x_t, y_t)$ and $p_t(x_t)$ are defined analogously for target domain.  
In feature-based unsupervised domain adaptation, our objective is to train a classifier 
$f_{\phi,\theta} = h_\phi \circ g_\theta$ which can perform well on target domain. Specifically, $g_\theta: \mathcal{X} \rightarrow \mathcal{Z}$ is the feature extractor, which is a projection function from data space $\mathcal{X}$ to latent feature space $\mathcal{Z}$, and $h_\phi: \mathcal{Z} \rightarrow \mathcal{P}(\mathcal{Y})$ is a classification function on the representation space, where $\mathcal{P}(\mathcal{Y})$ denotes the set of probability distributions over the label set $\mathcal{Y}$.
To address the covariate shift problem, many domain adaption methods are proposed to minimize the following objective motivated by the theory in \cite{ben2010theory,ganin2016domain}:
\begin{align}
 \label{domain_adt}
 \mathbb{E}_{(x_s,y_s)\sim p_s} \mathcal{L}_c (f_{\phi,\theta}(x_s),y_s)+ \lambda d(q_s(z_s; \theta),q_t(z_t; \theta))
\end{align}
Here $z_s$ and $z_t$ are latent representations for source and target domain; $q_s$ and $q_t$ are the marginal distributions of $z_s$ and $z_t$, which is implicitly defined by the marginals $p_s(x_s), p_t(x_t)$ and the deterministic mapping $g_\theta$; $\mathcal{L}_c$ is the cross entropy loss for training a classifier;
$d(\cdot, \cdot)$ is some divergence or distance measure between two distributions and $\lambda$ is the weighting factor. For instance, in \cite{long2015learning}, the divergence is realized as maximum mean discrepancy (MMD) and in many adversarial domain adaptation methods \cite{hoffman2017cycada,ganin2016domain,luo2017label}, the Jensen-Shannon divergence between $q_s$ and $q_t$ is minimized within an adversarial learning framework \cite{goodfellow2014generative}:
{\small
\begin{align}
\min_\theta \max_\omega \mathbb{E}_{x_s\sim p_s}  \log D_\omega (g_\theta(x_s))+ \mathbb{E}_{x_t\sim p_t}\log(1-  D_\omega(g_\theta(x_t))) \nonumber
\end{align}
}
where $D_\omega$
is a domain classifier on the representation space. Intuitively, the domain classifier is trained to distinguish the latent representations of source domain from that of target domain, while the feature extractor is jointly trained to confuse the discriminator by maximizing its classification error. At optimality, the marginal distributions of latent representations will be matched and the learned representations will be domain-invariant.

\subsection{Information Bottleneck Principle}\label{sec:information-bottleneck}
Let random variable $X$ denote the original signal and random variable $Y$ denote an output variable (\emph{e.g.} desired label), whose information we want to preserve. Given their joint distribution $p(X,Y)$, assuming the statistical dependence between $Y$ and $X$, the mutual information $I(X;Y)$ measures the mutual dependence between these two random variables. In this case, $Y$ implicitly determines both the relevant and irrelevant features in $X$. The information bottleneck(IB) method seeks to find an optimal representation of $X$ which captures the relevant part and filters out the irrelevant part. 

Formally, in the context of information bottleneck,
we are interested in finding the relevant part of $X$ with respect to $Y$, denoted by $Z$, the \emph{minimal sufficient statistics} of $X$ with respect to $Y$. Thus we assume the following \emph{Markov chain}: $Y \rightarrow X \rightarrow Z$ and we can obtain the optimal representation by minimizing $I(X;Z)$ under a constraint on $I(Y; Z)$ (to ensure the predictive ability of $Z$).

The objective of finding the optimal representation can be further formulated as the maximization of the following Lagrangian \cite{tishby2015deep}:
\begin{align}
    \label{eq:IB_lag}
    L(p(z|x)) =  I(Y;Z) -\beta I(X;Z)
\end{align}
subject to the Markov chain constraint. Here the positive Lagrangian multiplier $\beta$ represents a tradeoff between the complexity of the representation ($I(X;Z)$) and the amount of preserved relevant information ($I(Y;Z)$).
In essence, information bottleneck principle explicitly enforces the learned representation $Z$ to only preserve the information in $X$ that is useful to the prediction of $Y$, \emph{i.e.}, the \emph{minimal sufficient statistics} of $X$ with respect to $Y$.

In this paper, under the framework of information bottleneck principle, we propose a novel domain adaptation method which enforces the feature extractor to focus on the relevant factors implicitly defined by the task, and provide a thorough analysis of the benefits brought by our method both emipirically and theoretically.
\section{Method}

\subsection{Motivations}
\label{sec:motivation}
\cite{ganin2016domain} claimed that a successful adaptation can be achieved when the source domain classification error and the domain confusion loss are both small, which can be realized through optimizing the objective in Equation~(\ref{domain_adt}).

From the perspective of information preference, we can reformulate the objective in Equation~(\ref{domain_adt}) and understand the weakness of the constraint in a more straightforward way. 
To begin with, we split the loss function in Equation~(\ref{domain_adt}) into two terms, $  \mathbb{E}_{(x_s,y_s)\sim p_s} \mathcal{L}_c (f_{\phi,\theta}(x_s),y_s)$ and $\lambda d(q_s(z_s; \theta),q_t(z_t; \theta))$. In the following, we will show that minimizing the first term is equivalent to maximizing a variational lower bound  of the mutual information between learned representations and the labels in source domain (\textit{i.e.}, $I(Y_s;Z_s)$), and minimizing the second term corresponds to finding the domain-invariant features. To see these, let us first rewrite the negative of cross entropy loss as:
\begin{align}
    &-\mathbb{E}_{(x_s,y_s)\sim p_s} \mathcal{L}_c(f_{\phi,\theta}(x_s),y_s) \nonumber\\
    =&\mathbb{E}_{(x_s,y_s)\sim p_s}\left[\int p_\theta(z_s|x_s)\log h_\phi(y_s|z_s)dz_s\,\right]\, \nonumber\\
    =&\int p_s(x_s, y_s) p_\theta(z_s|x_s) \log h_\phi(y_s|z_s) dx_s \, dy_s \, dz_s\label{eq:cross-entropy-inital}
\end{align}
where $p_\theta(z_s|x_s)$ denotes the conditional distribution implied by the projection function $g_\theta$ (when $g_\theta$ is a deterministic projection, $p_\theta(z_s|x_s)$ corresponds to a delta distribution with non-zero density at $z=g_\theta(x_s)$). With the Markov chain assumption introduced in the Information Bottleneck Principle section 
, Equation~(\ref{eq:cross-entropy-inital}) can be rewritten as:
\begin{align}
    &\int p_s(x_s, y_s) p_\theta(z_s|x_s, y_s) \log h_\phi(y_s|z_s) dx_s \, dy_s \, dz_s \nonumber\\
    =&\int p_\theta(x_s, y_s, z_s) \log h_\phi(y_s|z_s) dx_s \, dy_s \, dz_s \nonumber \\
    =&\int p_\theta(y_s, z_s) \log h_\phi(y_s|z_s) dy_s \, dz_s \nonumber\\
    \leq& \int p_\theta(y_s, z_s) \log p_\theta(y_s|z_s) dy_s \, dz_s = I(Y_s;Z_s) - H(Y_s) \nonumber
\end{align}
Here, the inequality holds for the fact that $\KL(p_\theta(y|z) \| h_\phi(y|z)) \geq 0$. Since $H(Y)$ is a constant in our optimization procedure of $\theta$ and $\phi$, we know that minimizing the first term in Equation~(\ref{domain_adt}) corresponds to maximizing a lower bound of $I(Y_s;Z_s)$.

The second term $\lambda d(q_s(z_s; \theta),q_t(z_t; \theta))$ accounts for matching the marginal distribution of latent variables under the guidance of some distance or divergence. One notable example is the optimization of Jensen-Shannon divergence with adversarial training. Essentially, this constraint seeks to find the domain-invariant features of $X$. However, it should be noted that matching the marginals of latent features is \emph{agnostic} to the task of interest, which implies that the preserved domain-invariant features is likely to contain factors that are irrelevant to the prediction of desired labels. From learning theory \cite{sontag1998vc}
, we know that when the sample size is finite, the irrelevant factors (for the predictive task) in the noisy inputs can decrease the generalization ability of the models. We provide a formal discussion about the generalization error bound in Theoretical Analysis section.
From this perspective, we know that one direction to improve domain adaptation models is to add more constraints on the representation space so that the preserved features will not only be domain-invariant, but also relevant to the task of interest. 

On the other hand, due to the supervised learning objective in source domain, the learned representation with Equation~(\ref{domain_adt}) will intrinsically tend to capture the relationship between data instances and labels from source domain, while taking less attention on target domain. To take the label information for target domain where the exact label is not available into account during the feature learning, the cluster assumption can be adapted \cite{RTN,chapelle2005semi}, where the input distribution is assumed to contain separated data clusters and that data samples in the same cluster share the same class label. Cluster assumption introduces an inductive bias where we are seeking decision boundaries that do not go through high-density regions, which can be implemented through the following conditional entropy minimization:
{\small
\begin{align}
 \label{eq:conditional_en}
\mathcal{L}_{ce} \!=\! \mathbb{E}_{x_t \sim p_t(x_t), z_t \sim p_\theta(z_t| x_t)} \! \left[ \int h_\phi(y_t|z_t)\log h_\phi(y_t|z_t) dy_t \right]
\end{align}}

Note that the cluster assumption is satisfied when the learned representations only preserve semantic information that is relevant to the predictive task, it is strongly motivated to find a clean representation space with information bottleneck to justify the use of strong assumptions in domain adaptation methods. 

\subsection{Variational Bottleneck Domain Adaption}
\label{vbda}
Inspired by conditional entropy minimization in semi-supervised learning
\cite{chapelle2005semi,grandvalet2005semi} and deep variational information bottleneck \cite{tishby2000information,alemi2016deep}, to achieve better generalization ability, we propose a new regularization mechanism for domain adaptation, which explicitly enforces the feature extractor to only preserve the minimal sufficient statistics of the input data with respect to the labels for both source and target domain.

As discussed in the Motivations section,
from the perspective of information bottleneck principle, we know that the objective in Equation~(\ref{domain_adt}) lacks a constraint for minimizing the mutual information between $X$ and $Z$:
\begin{align}
     I_\theta(X;Z) =  \int p_\theta(x,z) \log \frac{ p_\theta(z|x)}{ p_\theta(z)} dx \, dz
\end{align}

However, it should be noted that in general, directly computing and optimizing $I(X;Z)$ is computationally intractable\cite{alemi2016deep}, as it requires solving an integral over latent feature space. To achieve tractability, we follow the methods proposed in \cite{alemi2016deep} and instead optimize a tractable variational upper bound:
\begin{align}
\label{information_up_bound}
 I_\theta(X;Z)
 &= \! \int  p(x)  p_\theta(z|x) \log
 \frac{p_\theta(z|x)}{p_\theta(z)} dx \, dz \nonumber\\
 &=\mathbb{E}_{x \sim p(x)} \KL(p_\theta(z|x)\|r(z)) - \KL(p_\theta(z)\|r(z)) \nonumber\\
 &\leq \mathbb{E}_{x \sim p(x)} \KL(p_\theta(z|x)\|r(z))\nonumber\\
 &\triangleq  I_{U}(X;Z).
\end{align}

Here, $r(z)$ is the prior distribution of latent features and $p_\theta(z)$ denotes the marginal distribution implied by $p(x)$ and conditional distribution $p(z|x)$, and the inequality holds for the fact that $\KL(p_\theta(z)\|r(z)) \geq 0$.

To incorporate the above variational information bottleneck, with abuse of notation, we introduce a stochastic feature extracting function $g_\theta: \mathcal{X} \rightarrow \mathcal{P}(\mathcal{Z})$, which maps a sample $x$ to a stochastic representation $z \sim g_\theta(z|x)$. Now we can add the following terms to the objective in order to enforce the feature extractor to only preserve task-relevant factors:
$$\mathbb{E}_{x_s\sim p_s}\KL(g_\theta(z|x_s) \| r(z)) + \mathbb{E}_{x\sim p_t}\KL(g_\theta(z|x_t) \| r(z))$$

In our experiments, the stochastic feature extracting function is realized as a Gaussian distribution $g_\theta(z|x) =  \mathcal{N}(z|g_\theta^{\mu}(x), g_\theta^{\Sigma}(x))$, where $g_\theta(x)$ outputs the mean $\mu$ and diagonal covariance matrix $\Sigma$ of $z$. When $r(z)$ allows for the computation of Kullback-Leibler divergence analytically, the upper bound in Equation~(\ref{information_up_bound}) can be easily optimized. Thus we choose $r(z)$ to be a standard normal distribution, $r(z) = \mathcal{N}(0,I)$. 
Note that although the objective here shares similar mathematical form  with the KL regularization term in Variational Autoencoder (VAE) \cite{kingma2013auto}, the motivation and interpretation of the objectives are related but different. As a generative model, VAE consists of a pre-determined prior $p(z)$ for the latent variables and a stochastic decoder $p(x|z)$ for reconstruction. The amortized encoder $q(z|x)$ is introduced as a variational approximation to the true posterior $p(z|x) = p(x|z)p(z)/p(x)$ and the resulting evidence lower bound (ELBO) works as a tractable lower bound for the log-likelihood objective.
While in the variational information bottleneck, the $r(z)$ is introduced to derive a tractable upper bound for minimizing the mutual information term. 
Note that the equality in Equation~(\ref{information_up_bound}) holds only when $p_\theta(z)=r(z)$. Therefore,
by choosing a simple realization of $r(z)$ such as standard normal distribution, we are also introducing an inductive bias of regularizing the marginal distribution of the learned representations (\emph{i.e.}, $p_\theta(z)$) to be as simple as possible. 

Putting things together, the final objective function in our framework can be written as: 
\begin{align}
    \mathcal{L}(\theta, \phi) =
    ~&\mathbb{E}_{(x_s,y_s) \sim p_s, z_s \sim g_\theta(z|x_s)} \mathcal{L}_c(h_\phi(z_s),y_s)+ \nonumber\\\nonumber
    &\lambda_d \cdot d(q_s(z_s; \theta),q_t(z_t; \theta))+
    \lambda_{ce} \cdot \mathcal{L}_{ce} +  \\ \nonumber
    &\lambda_{s} \cdot \mathbb{E}_{x_s\sim p_s}\KL(g_\theta(z|x_s) \| r(z))+ \\ 
     &\lambda_{t} \cdot \mathbb{E}_{x_t\sim p_t}\KL(g_\theta(z|x_t) \| r(z))
\end{align}
Here, $\mathcal{L}_c$ is the classification loss; 
$q_s(z_s)$ and $q_t(z_t)$ are implicit marginal distributions induced by the marginal distributions $p_s(x_s), p_t(x_t)$ and the conditional distribution $g_\theta(z|x)$;
$\mathcal{L}_{ce}$ is the conditional entropy term defined in Equation~(\ref{eq:conditional_en}); $\lambda_d, \lambda_{ce}, \lambda_s$ and $\lambda_t$ are hyperparameters controlling the optimization tradeoff among each term.
Note that there is a stochastic structure in the model, we utilize the reparameterization trick introduced in \cite{kingma2013auto} to back-propagate unbiased estimated gradients through single example.

\subsection{Theoretical Analysis}
\label{domain adapt theory}
In this section, we analyze the theoretical properties of our proposed method.
\begin{theorem}[\cite{ben2010theory}]\label{the:1}
     Let $\mathcal{H}$ be the hypothesis space, Given $(X_s, \epsilon_s)$ and $(X_t, \epsilon_t)$ as the two domains and their corresponding test error functions. Then for any $h \in \mathcal{H}$, we have:
    \begin{align*}
    \epsilon_t(h) \le \frac{1}{2} d_{\mathcal{H}\Delta\mathcal{H}}(X_s, X_t) + \epsilon_s(h) + \min_{h' \in \mathcal{H}} \epsilon_t(h') + \epsilon_s(h')
    \end{align*}
\end{theorem}
Here $d_{\mathcal{H}\Delta\mathcal{H}}$ represents a discrepancy measure between source and target domain with respect to a hypothesis space $\mathcal{H}$, which is defined as:
\begin{align}\label{eq:adaptation_complexity}
    &d_{\mathcal{H}\Delta\mathcal{H}}(X_s, X_t)= \\\nonumber
    & 2 \sup_{h, h' \in \mathcal{H}}
    \| \mathbb{E}_{x \sim X_s} \left[h(x) \neq h'(x)\right] -
    \mathbb{E}_{x \sim X_t} \left[h(x) \neq h'(x)\right] \| .
\end{align}
For a fixed hypothesis space $\mathcal{H}$, $d_{\mathcal{H}\Delta\mathcal{H}}(X_s, X_t)$ is the intrinsic difference between source and target domain, which is fixed and determined by the characteristics of the data distributions.
Now we will show that how the $I(X;Z)$ term from information bottleneck principle can help minimize the test error term, \emph{i.e.} $\epsilon_s(h)$ and $\min_{h' \in \mathcal{H}} \epsilon_t(h') + \epsilon_s(h')$ in Theorem~\ref{the:1}.

\begin{theorem}[\cite{shamir2010learning}]
    \label{GEENRAL}
    For any probability distribution $p(x, y)$, with a probability of at least $1 - \delta$ over the draw of the sample of size $m$ from $p(x, y)$, $\hat{I} (X; Z )$ and $\hat{I} (Y ; Z )$ are the empirical estimate of the mutual information $I(X;Z)$ and $I(Y;Z)$. Then for any $Z$,
    \begin{align}\label{eq:gen-bound}
        &|I(Y;Z)-\hat{I}(Y;Z)| \leq \nonumber\\
        &\sqrt{\dfrac{C\log(|\mathcal{Y}|/\delta)}{m}}(C_1\log(m)\sqrt{|\mathcal{Z}I(X;Z)|} \nonumber\\
        &+C_2|\mathcal{Z}|^{3/4}(I(X;Z))^{1/4}+C_3\hat{I}(X;Z))
    \end{align}
where $C, C_1, C_2$ and $C_3$ are constants. $|\mathcal{Z}|$ and $|\mathcal{Y}|$ correspond to the cardinality
of variables $Z$ and $Y$.
\end{theorem}

\begin{table*}[htbp]
  \vspace{-3pt}
  \addtolength{\tabcolsep}{2pt}
  \centering
  \caption{Classification accuracy (\%) on {Office-31} (ResNet50)}
  \label{table:office31}
  \begin{tabular}{cccccccc}
    \toprule
    Method & A $\rightarrow$ W & D $\rightarrow$ W & W $\rightarrow$ D & A $\rightarrow$ D & D $\rightarrow$ A & W $\rightarrow$ A & Avg \\
    \midrule

   	ResNet-50 \cite{he2016deep} & 68.4$\pm$0.2 & 96.7$\pm$0.1 & 99.3$\pm$0.1 & 68.9$\pm$0.2 & 62.5$\pm$0.3 & 60.7$\pm$0.3 & 76.1 \\
    RTN \cite{RTN} & 84.5$\pm$0.2 & 96.8$\pm$0.1 & 99.4$\pm$0.1 & 77.5$\pm$0.3 & 66.2$\pm$0.2 & 64.8$\pm$0.3 & 81.6 \\
    DANN \cite{ganin2016domain} & 82.0$\pm$0.4 & 96.9$\pm$0.2 & 99.1$\pm$0.1 & 79.7$\pm$0.4 & 68.2$\pm$0.4 & 67.4$\pm$0.5 & 82.2 \\
    ADDA \cite{tzeng2017adversarial} & 86.2$\pm$0.5 & 96.2$\pm$0.3 & 98.4$\pm$0.3 & 77.8$\pm$0.3 & 69.5$\pm$0.4 & 68.9$\pm$0.5 & 82.9 \\
    MADA\cite{pei2018multi} & 90.0$\pm$ 0.1 &97.4 $\pm$ 0.1 &99.6 $\pm$ 0.1 &87.8 $\pm$ 0.2 &70.3 $\pm$ 0.3 &66.4 $\pm$ 0.3 &85.2 \\
    SimNet\cite{pinheiro2018unsupervised} &88.6$\pm$0.5 &98.2 $\pm$ 0.2 &99.7$\pm$0.2 &85.3 $\pm$ 0.3 & \textbf{73.4} $\pm$ 0.8&\textbf{71.6} $\pm$ 0.6  &86.2 \\
    GTA \cite{GTA} & 89.5$\pm$0.5 & 97.9$\pm$0.3 & 99.8$\pm$0.4 & 87.7$\pm$0.5 & 72.8$\pm$0.3 & 71.4 $\pm$0.4 & 86.5 \\
    \textbf{VBDA} & \textbf{92.1}$\pm$0.1 & \textbf{98.6}$\pm$0.1 & \textbf{100.0}$\pm$.0 & \textbf{93.2}$\pm$0.2 & 69.4$\pm$0.1 & {69.1}$\pm$0.3 & \textbf{87.0} \\
    \bottomrule
  \end{tabular}
\end{table*}

\begin{table*}[htbp]
  \vspace{-3pt}
  \addtolength{\tabcolsep}{-3pt}
  \centering
  \caption{Accuracy (\%) on {Office-Home} for unsupervised domain adaptation (ResNet50)}
  \label{table:officehome}
  \begin{tabular}{cccccccccccccc}
    \toprule
    Method & Ar$\shortrightarrow$Cl & Ar$\shortrightarrow$Pr & Ar$\shortrightarrow$Rw & Cl$\shortrightarrow$Ar & Cl$\shortrightarrow$Pr & Cl$\shortrightarrow$Rw & Pr$\shortrightarrow$Ar & Pr$\shortrightarrow$Cl & Pr$\shortrightarrow$Rw & Rw$\shortrightarrow$Ar & Rw$\shortrightarrow$Cl & Rw$\shortrightarrow$Pr & Avg \\
    \midrule
     ResNet-50 \cite{he2016deep} & 34.9 & 50.0 & 58.0 & 37.4 & 41.9 & 46.2 & 38.5 & 31.2 & 60.4 & 53.9 & 41.2 & 59.9 & 46.1 \\
    DAN \cite{long2015learning} & 43.6 & 57.0 & 67.9 & 45.8 & 56.5 & 60.4 & 44.0 & 43.6 & 67.7 & 63.1 & 51.5 & 74.3 & 56.3 \\
    DANN \cite{ganin2016domain} & 45.6 & 59.3 & 70.1 & 47.0 & 58.5 & 60.9 & 46.1 & 43.7 & 68.5 & 63.2 & 51.8 & 76.8 & 57.6 \\
    JAN \cite{JAN} & 45.9 & 61.2 & 68.9 & 50.4 & 59.7 & 61.0 & 45.8 & 43.4 & 70.3 & 63.9 & 52.4 & 76.8 & 58.3 \\
    CDAN \cite{CDAN} & \textbf{49.0} & {69.3} & 74.5 & 54.4 & {66.0} & {68.4} & {55.6} & \textbf{48.3} & 75.9 & 68.4 & \textbf{55.4} & {80.5} & 63.8 \\
    \textbf{VBDA} & 45.6 & \textbf{70.7} & \textbf{75.0} & \textbf{58.1} & \textbf{70.0} & \textbf{68.8} & \textbf{56.1} & 45.8 & \textbf{76.2} & \textbf{69.1} & 53.8 & \textbf{81.3} & \textbf{64.21} \\
    \bottomrule
  \end{tabular}%
\end{table*}

Theorem.~\ref{GEENRAL} shows that the $|I(Y;Z)-\hat{I}(Y;Z)|$ which is a measure of difference between training and test error is bounded by a monotonic function of $I(X;Z)$.  Essentially, it is true that minimizing $I(X;Z)$ will minimize the generalization error, but this is not enough. A degenerate case is $I(X;Z) = 0$, in which case the prediction is random, although the difference between training and test error is zero. So we also need to make sure both the training error and the generalization error is small. We can decrease $\epsilon_s(h)$ with information bottleneck (IB) principle, since we are explicitly minimizing the training error in source domain and the generalization error in both domains. For $\min_{h' \in \mathcal{H}} \epsilon_t(h') + \epsilon_s(h')$, ideally IB will not harm predictive ability by just removing irrelevant factors, so the combined training error $\min_{h' \in \mathcal{H}} \hat{\epsilon_t}(h') + \hat{\epsilon_s}(h')$ should be the same with or without IB. While the combined test error is the sum of combined training error and combined generalization error, we are also able to reduce the combined test error.

\begin{table}
\centering
\caption{Classification accuracies (\%) on digits datasets.}
\vspace{-3pt}
\begin{tabular}{c|ccc}
\toprule
Source Domain   & M & U & S\\
Target Domain & U & M & M\\
\midrule
UNIT\cite{liu2017unsupervised} & \textbf{96.0} & 93.6 & 90.5\\
CyCADA \cite{hoffman2017cycada} &  95.6 & 96.5 & 90.4 \\

RAAN \cite{RAAN} & 89.0 & 92.1 & 89.2 \\
CDAN \cite{CDAN} & 95.6 & \textbf{98.0} & 89.2 \\
\textbf{VBDA}(ours) & \textbf{96.0}&\textbf{98.0}&\textbf{93.8}\\
\bottomrule
\end{tabular}
\label{table:digits}
\end{table}

\section{Experiments}
We conduct  experiments on various visual domain adaptation benchmarks including \textbf{Office-31}, \textbf{Office-home} and \textbf{Digits}, to compare our approach against state-of-the-art deep domain adaptation methods. 
\subsection{Setup}

\textbf{Office-31} \cite{office31}  is a widely-used dataset for visual domain adaptation, with 4,652 images and 31 categories from three distinct domains: Amazon ($\textbf{A}$), which contains images downloaded from amazon.com, Webcam ($\textbf{W}$) and DSLR ($\textbf{D}$), which contain images taken by web camera and digital SLR camera respectively. We denote the three domains as $\textbf{A}$, $\textbf{W}$ and $\textbf{D}$. By permuting the 3 domains, we get 6 domain adaptation tasks.

\textbf{Office-home} \cite{officehome}
is a better organized and more difficult dataset than Office-31, which consists of 15,500 images in 65 object classes in office and home settings. It consists of four extremely dissimilar domains: Artistic images ($\textbf{Ar}$), Clip Art ($\textbf{Cl}$), Product images ($\textbf{Pr}$), and Real-World images ($\textbf{Rw}$). There are 12 domain adaptation tasks by permuting the 4 domains.

\textbf{Digits} We also explore three digits datasets of varying difficulty, $\textbf{MNIST}$, $\textbf{SVHN}$ and $\textbf{USPS}$. Following the evaluation protocol of CyCADA \cite{hoffman2017cycada}, we investigate the following three tasks: USPS to MNIST ($\textbf{U}\rightarrow \textbf{M}$), MNIST to USPS ($\textbf{M}\rightarrow \textbf{U}$) and SVHN to MNIST ($\textbf{S} \rightarrow \textbf{M}$).

We follow the standard protocols for evaluating unsupervised domain adaptation \cite{long2015learning,ganin2016domain}. In the experiments, we  observed that the hyperparameters ($\lambda_d$, $\lambda_s$, $\lambda_t$, $\lambda_{ce}$) are easy to choose and work well across multiple tasks. Specifically, we keep a fixed weight $\lambda_d$ for domain adversarial loss and we choose the value of $\lambda_s$, $\lambda_t$, $\lambda_{ce}$ from a small candidate set, \emph{i.e.}, $\{0.1,0.01\}$.
The hyperparameters $\lambda_s$,$\lambda_t$ for variational information bottleneck are selected according to the entropy of domain. For example, the higher-entropy domain tends to hold more irrelevant information and needs a larger mutual information regularization weight. The experiments on \textbf{Office-31} and \textbf{Office-home} is implemented based on ResNet-50 \cite{he2016deep} pretrained on the ImageNet dataset \cite{deng2009imagenet}. As for the digits dataset, we train our models with a small CNN \cite{french2017self}.

\begin{figure*}[ht]
  \centering
    \subfigure[Test accuracy for different $\lambda_s$.($\lambda_t=0$)]{
    \includegraphics[width=0.42\textwidth]{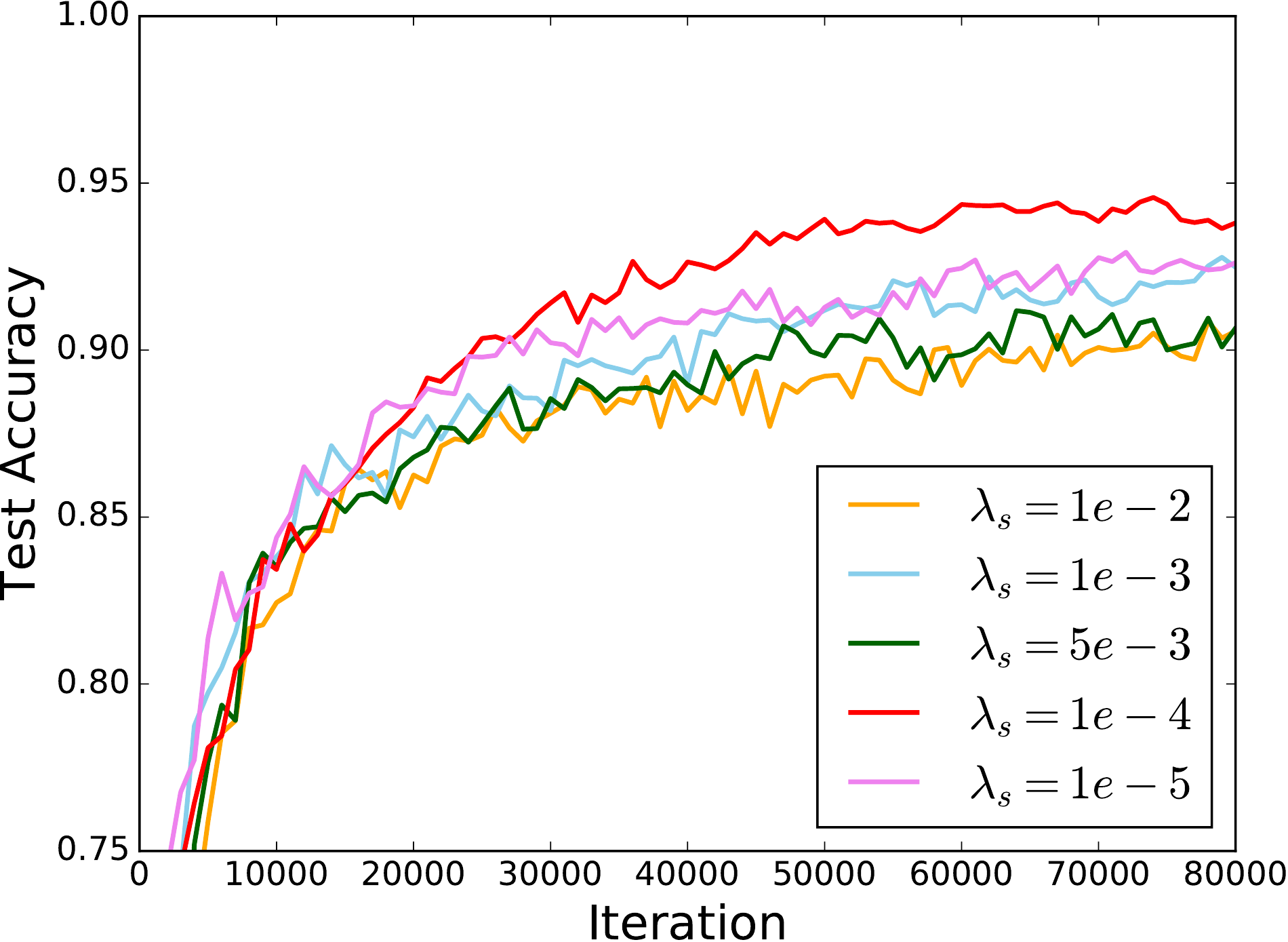}
    \label{hyperpara}
  }\hfil
  \subfigure[Test accuracy for different $\lambda_t$.($\lambda_s=1e-4$)]{
    \includegraphics[width=0.42\textwidth]{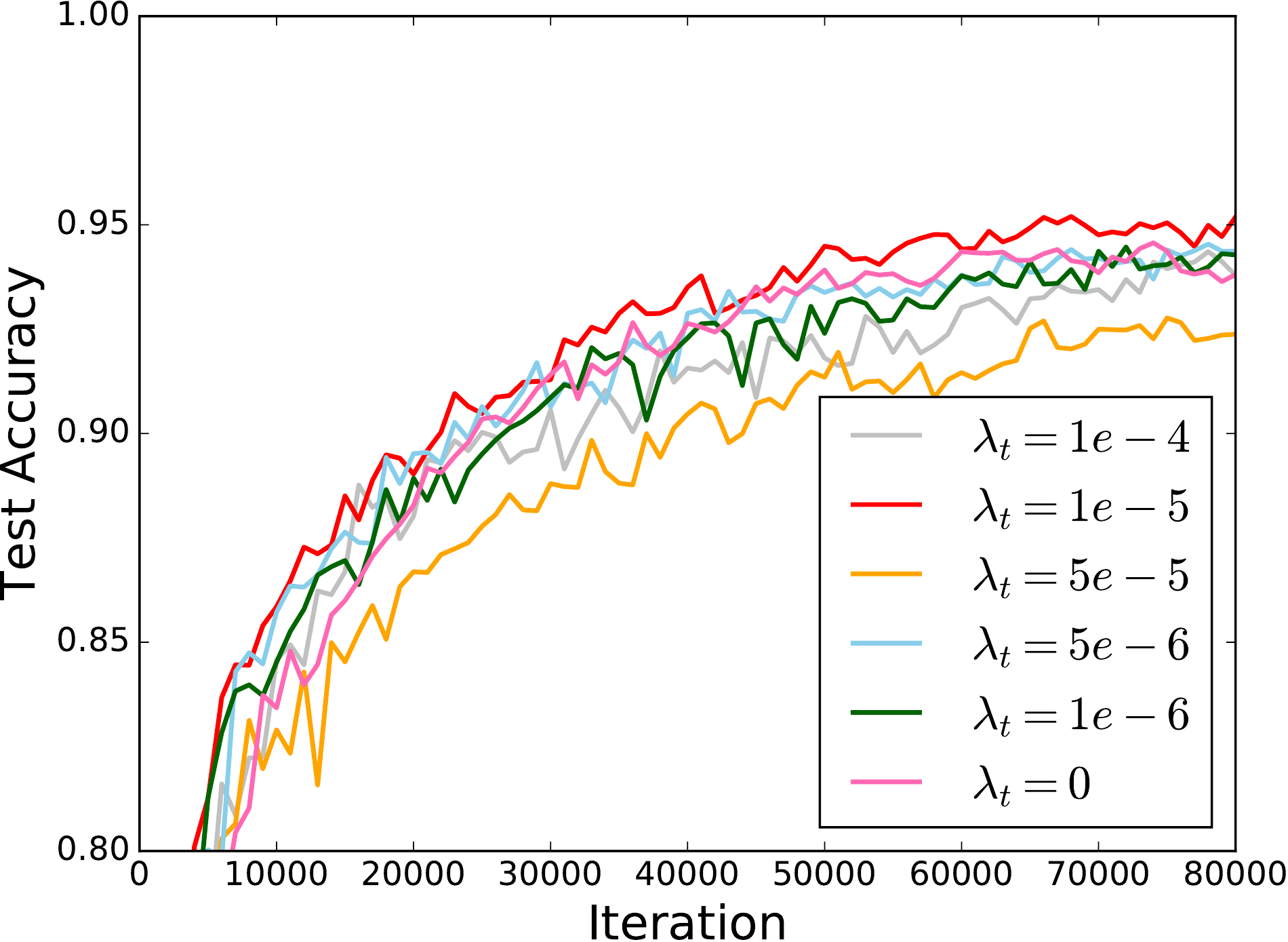}
    \label{hyper_t}
  }\hfil
  \subfigure[ DANN+CE vs VBDA]{
    \includegraphics[width=0.42\textwidth]{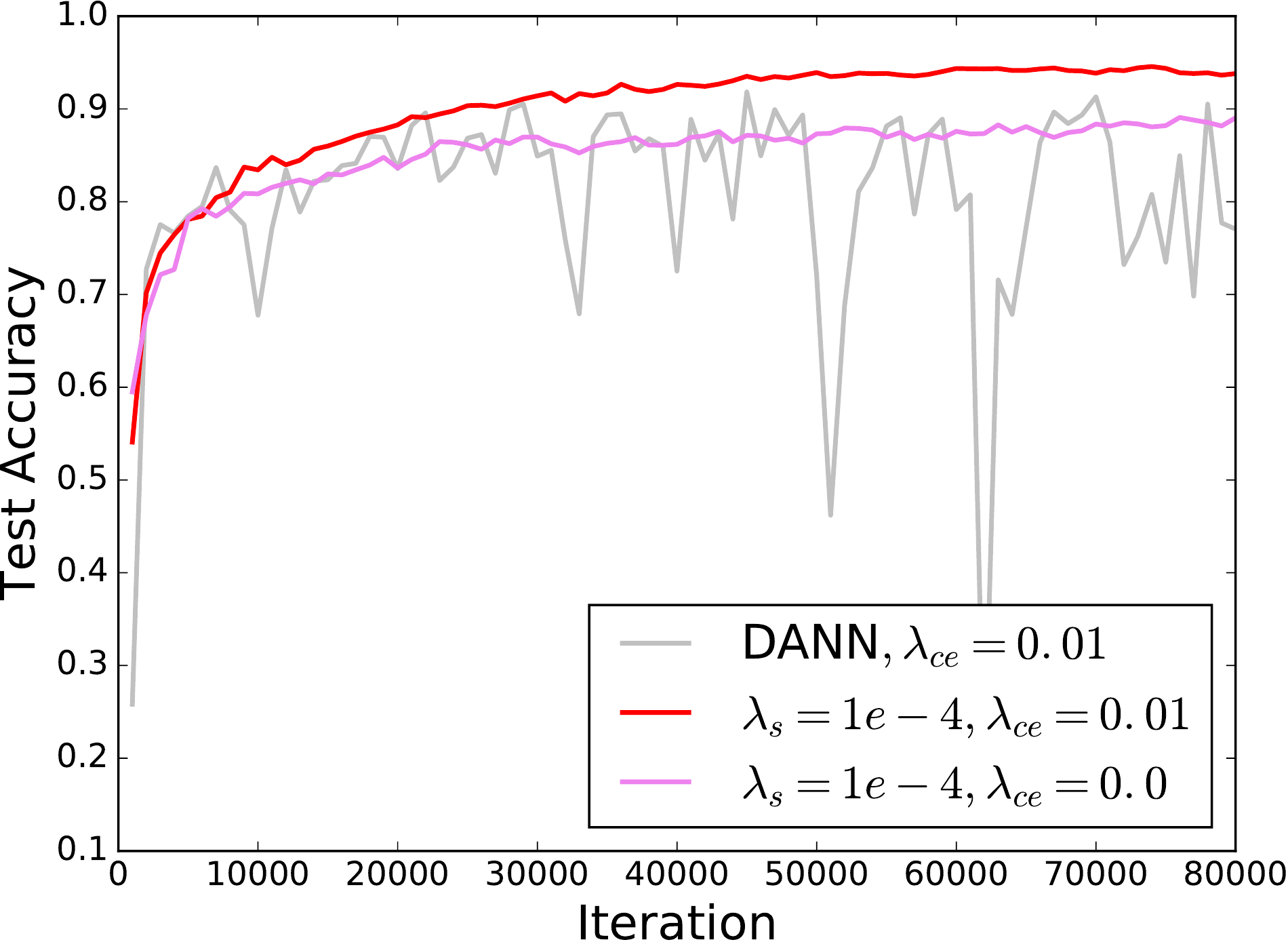}
    \label{trainng_curve}
  }\hfil
  \subfigure[$I_U(X;Z)$ and $I_L(Y;Z)$]{
    \includegraphics[width=0.42\textwidth]{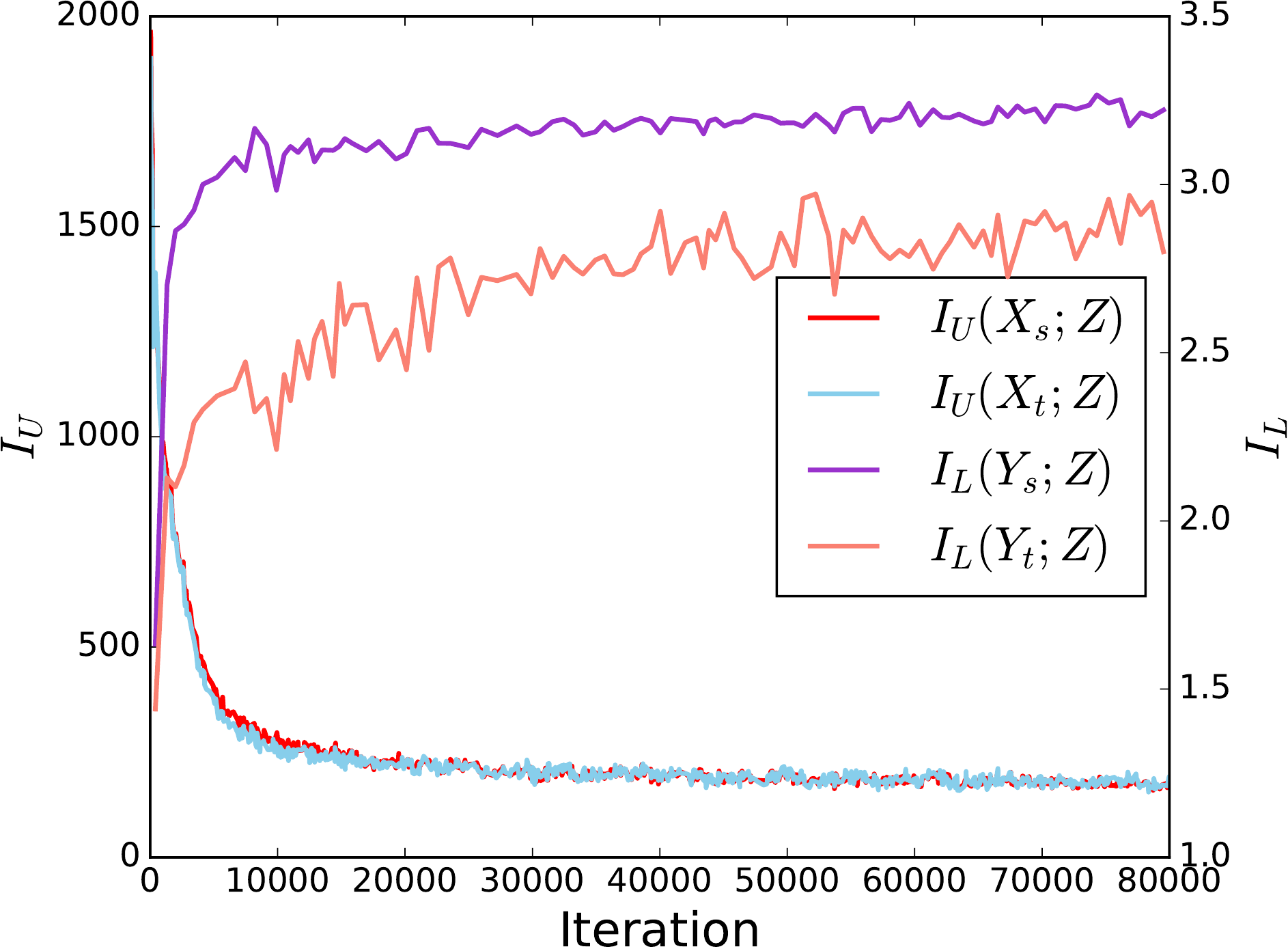}
    \label{iucurve}
  }
  \vspace{-10pt}
  \caption{The sensitivity of the accuracy w.r.t the value of $t_h$ (left) and $t_l$ (right).}
  \label{fig:analysis}
    \vspace{-10pt}
\end{figure*}


\begin{figure*}[ht]
\centering
\subfigure[Resnet]{\includegraphics[width=0.45\textwidth]{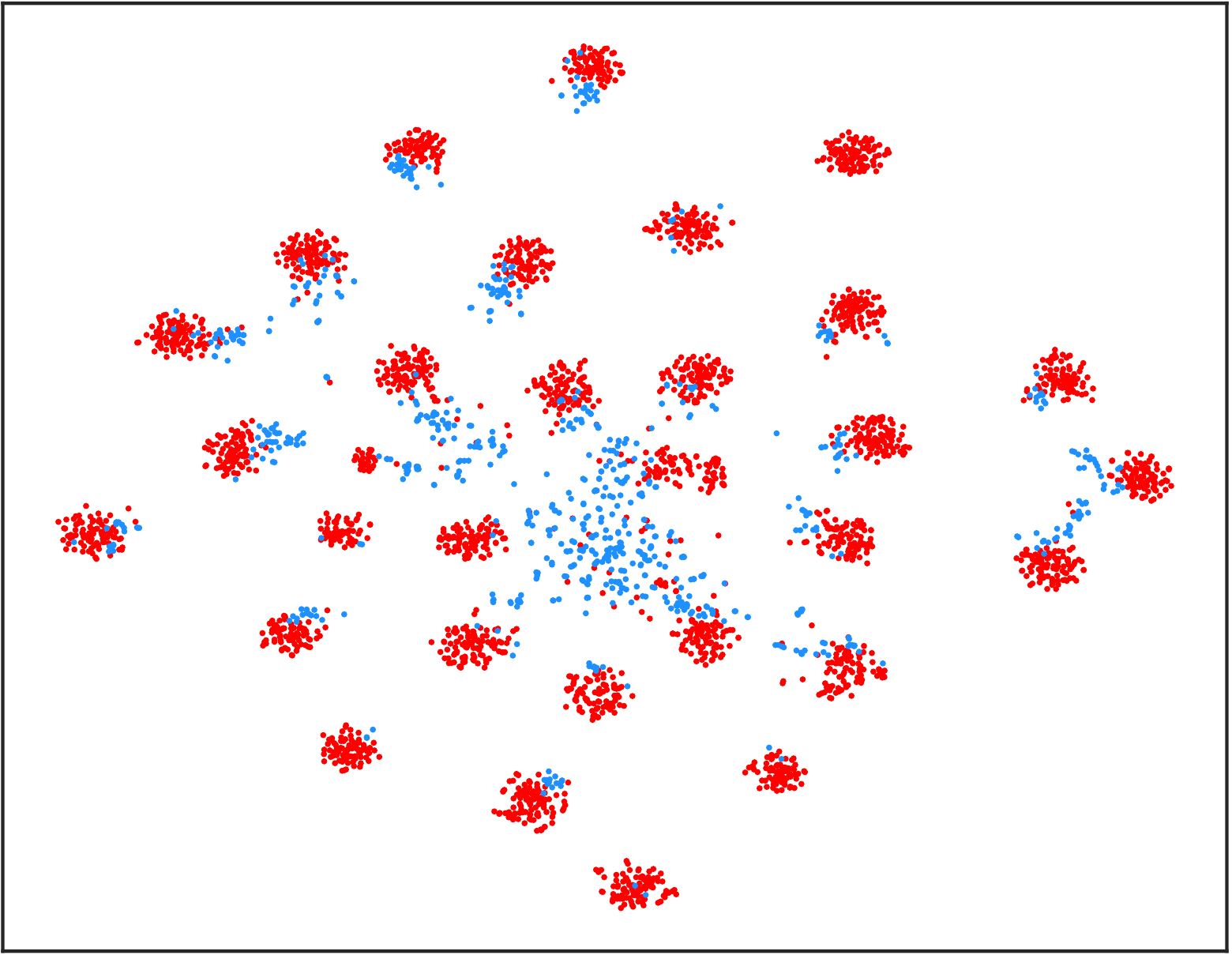}\label{tsne-resent}}\hfil
\subfigure[DANN]{\includegraphics[width=0.45\textwidth]{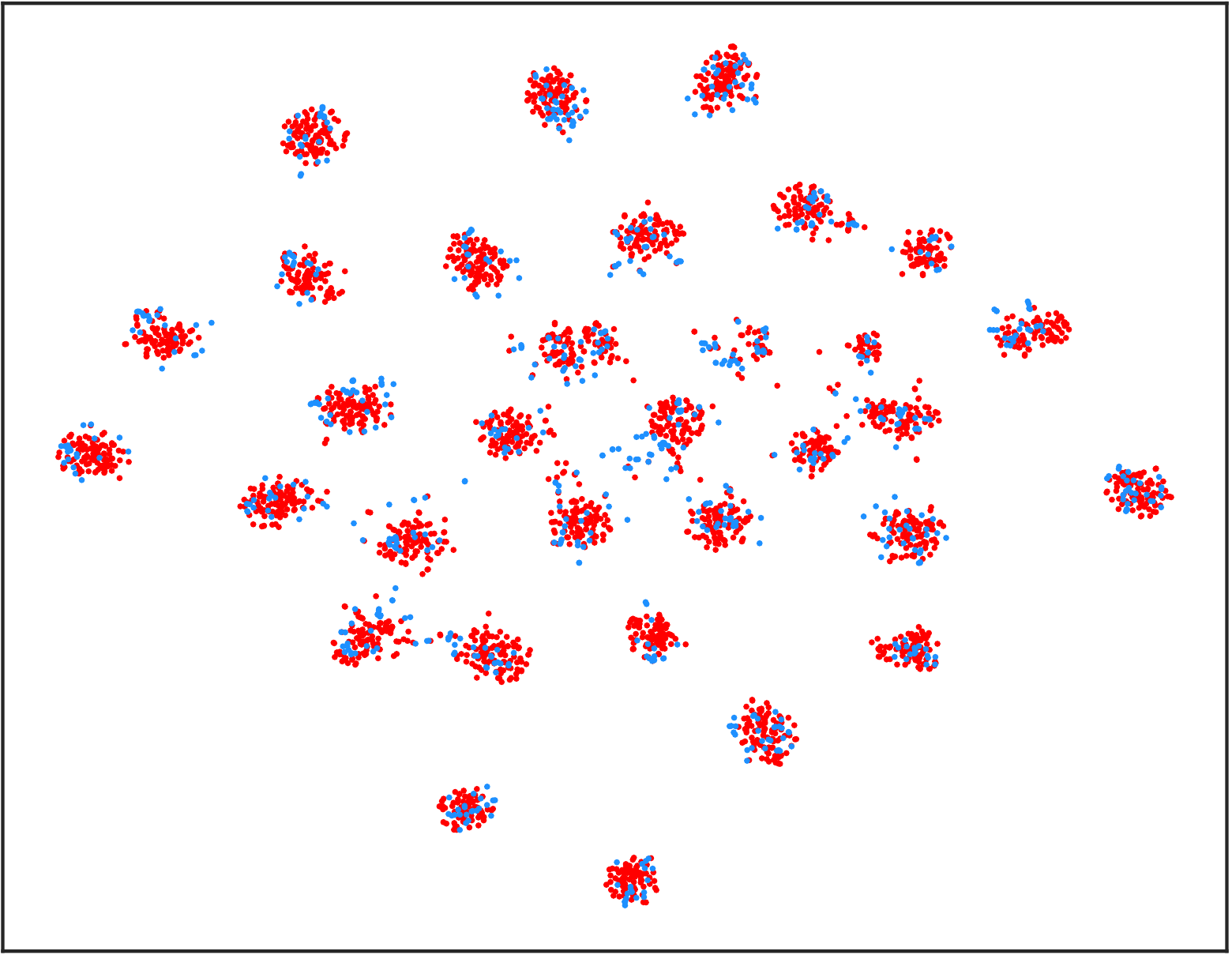}\label{tsne-dann}}\hfil
\subfigure[MADA]{\includegraphics[width=0.45\textwidth]{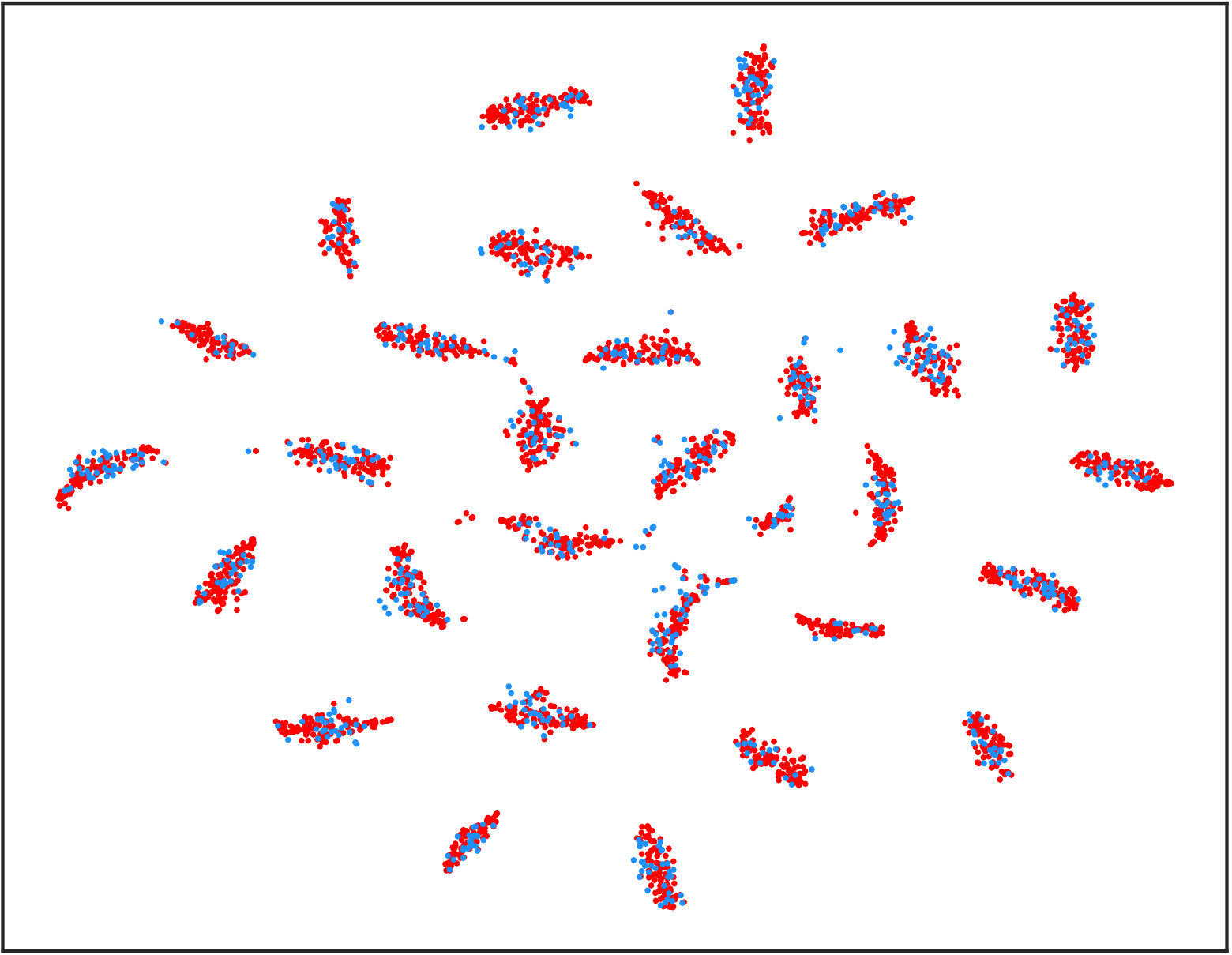}\label{tsne-cdan}}\hfil
\subfigure[VBDA]{\includegraphics[width=0.45\textwidth]{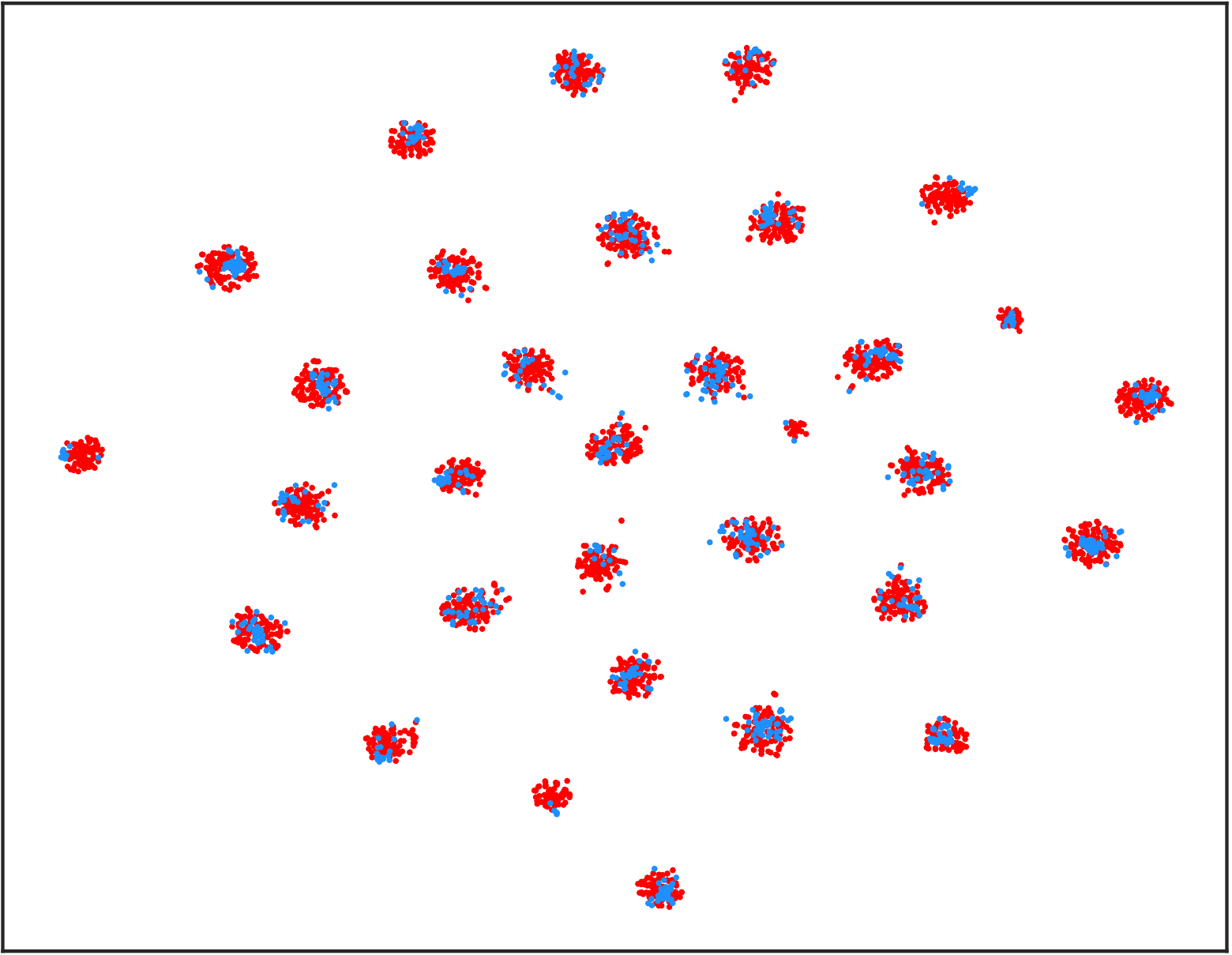}\label{tsne-vbda}}
\caption{The t-SNE visualization of Resnet, DANN, MADA and VBDA on task $\textbf{A} \rightarrow \textbf{W}$ (red:$\textbf{A}$,blue:$\textbf{W}$) }
\label{tsne}
\end{figure*}

\subsection{Results}
The results on the Office-31 dataset are reported in Table \ref{table:office31}. For fair comparison, the baselines are directly reported from their original papers if the protocol is the same. Our VBDA model remarkably outperforms all comparison methods on most of the tasks. Notably, the model performance are remarkably improved on the hard task, \textit{e.g.}, $A\rightarrow W$, $A \rightarrow D$, where the two domain are significantly different. The interpretation follows that the variations
of the source and target domain in these tasks are substantially different, and the task-irrelevant information are the main obstacles for adapting model. Thus this demonstrate that VBDA is good at eliminating these factors and focusing on the essential information for the task of interest. The performance is also further promoted on the relatively easy tasks, such as $D \rightarrow W$ and $W \rightarrow D$. However, the model performance on the tasks, $W \rightarrow A$ and $D \rightarrow A$ are slightly lower than some approaches. This is due to the fact that, the average number of images for 31 classes in  Webcam and DSLR are only 26 and 16, which are much lower than the number of bins for representing the image distribution and make the empirically estimated mutual information bounds not reliable enough for applying effective information bottleneck.

The results on the Office-home can be found in Table \ref{table:officehome}. The VBDA method significantly promotes the accuracy on most domain adaptation tasks and outperforms CDAN, a state-of-the-art method on this dataset by 0.41\% on average. The Office-home is a more challenging dataset, which has four domains with larger domain gap and more categories. The information difference between the four domains are more obvious, \textit{i.e.} \textbf{Rw} and \textbf{Ar} contains much more redundant information than \textbf{Cl} and \textbf{Pr} for classification task, and the information bottleneck can help control the information flow flexibly to learn \textit{clean} representation for adaptation and classification. The desirable performance on such challenging domain adaption tasks highlights the effectiveness of matching essential information by utilizing information bottleneck principle.

The results on digits datasets are shown in Table~(\ref{table:digits}). In task MNIST$\rightarrow$USPS and USPS$\rightarrow$MNIST, VBDA performs better or at least comparably with previous methods. And on the more challenging setting, SVHN$\rightarrow$MNIST, our model promotes the existing methods by 3.3\%. In particular, VBDA outperforms CyCADA, a state-of-the-art pixel-level adaptation method, which further proves the efficacy of VBDA.

\subsection{Analysis and Ablation Study}
To make a distinction between the utility of two main components of VBDA: the conditional entropy term and the information bottleneck term, we conduct a case study on task SVHN$\rightarrow$MNIST. We can observe that in Fig~(\ref{trainng_curve}), with conditional entropy term only(DANN+CE), the training is quite unstable; with bottleneck term only, the training is stable while the performance declines; with both terms, the model converges stably to a best test accuracy on target domain. 

We also conduct ablation studies on hyper-parameter learning for $\lambda_s$ and $\lambda_t$ in task SVHN$\rightarrow$MNIST. $\lambda_t$ is preferred to be smaller than $\lambda_s$, since SVHN has more irrelevant information to be penalized than MNIST. From Fig~(\ref{hyperpara}), we can observe, the accuracy suffers with a too large $\lambda_s$. As $\lambda_s$ becomes larger, we forget more about the input and the learned representation start to become more and more indistinguishable. And the best performance is achieved with an intermediate value of $\lambda_s$, in this case, the best setting is $\lambda_s=1e-4$. Similar phenomenon can be observed on the $\lambda_t$ in Fig~(\ref{hyper_t}).

And the mutual information changes during the optimization is showed in the Fig~(\ref{iucurve}). As we can observe,  the mutual information between the representation and label, \textit{i.e.}, $I(Y_s;Z)$ and $I(Y_t;Z)$,  are both improved during the training and the mutual information upper bound, \textit{i.e.}, $I_U(X_s;Z)$ and $I_U(X_t;Z)$, between input and representation gradually declined, which indicates that more semantic information has been embedded and more nuisances have been removed in the representation space.

\subsection{Feature Visualization}
The t-SNE visualization of representation in task A$\rightarrow$W (31 classes) is illustrated in Fig~(\ref{tsne}). Note that the source and target representation is not aligned well by Resnet. DANN can match the marginal feature distribution, but there are still target points near or across the class boundary. MADA aligns the source and target domain and discriminates categories better, but each class are more scattered than that in VBDA and some target points deviate from the corresponding cluster center. VBDA has clearer cluster boundary and more compact and centered clusters, demonstrating that information irrelevant to classification is filtered by the proposed variational information bottleneck and only information relevant to classification is preserved.

\section{Conclusions}
In this paper, we proposed Variation  Bottleneck  Domain
Adaptation (VBDA), a simple yet effective regularization mechanism for unsupervised domain adaptation. VBDA enhances semantic information and removes irrelevant factors in the learned representation space, which improves generalization ability and renders strong hypothesis such as cluster assumption more realistic. Comprehensive experiments demonstrate that the proposed approach achieves state-of-the-art performance on various domain adaptation benchmarks.

\bibliographystyle{ecai}
\bibliography{reference}

\end{document}